\documentclass[10pt]{article}

\usepackage{balance} 
\usepackage[english]{babel}
\usepackage[utf8]{inputenc}
\usepackage{subcaption}
\usepackage{multirow}
\usepackage{multicol}
\usepackage{amsmath}
\usepackage{mathtools}

\usepackage{enumitem}
\usepackage{siunitx}
\usepackage{graphicx}
\usepackage{fullpage}
\usepackage{natbib}
\usepackage{booktabs}
\usepackage{url}
\usepackage{subcaption}
\usepackage{amsfonts}

\title{Multiclass Common Spatial Pattern for EEG based Brain Computer Interface with Adaptive Learning Classifier}

\author{Hardik Meisheri$^1$, Nagaraj Ramrao$^1$, Suman Mitra$^1$\\$^1$DA-IICT, Gandhinagar, India}

\newcommand{\BibTeX}{\rm B\kern-.05em{\sc i\kern-.025em b}\kern-.08em\TeX}

\begin{document}

	\date{}
	\maketitle 
	
	\begin{abstract}
		
		In Brain Computer Interface (BCI), data generated from Electroencephalogram (EEG) is non-stationary with low signal to noise ratio and contaminated with artifacts. Common Spatial Pattern (CSP) algorithm has been proved to be effective in BCI for extracting features in motor imagery tasks, but it is prone to overfitting. Many algorithms have been devised to regularize CSP for two class problem, however they have not been effective when applied to multiclass CSP. Outliers present in data affect extracted CSP features and reduces performance of the system. In addition to this non-stationarity present in the features extracted from the CSP present a challenge in classification. We propose a method to identify and remove artifact present in the data during pre-processing stage, this helps in calculating eigenvectors which in turn generates better CSP features. To handle the non-stationarity, Self-Regulated Interval Type-2 Neuro-Fuzzy Inference System (SRIT2NFIS) was proposed in the literature for two class EEG classification problem. This paper extends the SRIT2NFIS to multiclass using Joint Approximate Diagonalization (JAD). The results on standard data set from BCI competition IV shows significant increase in the accuracies from the current state of the art methods for multiclass classification.
		
	\end{abstract}
	
	
	\section{Introduction}
	
	There is on-going research to connect brain signals with computers/machines in order to create a comfortable communication pathway for disabled people with external environment. Brain Computer Interface (BCI) aims to provide non-muscular communication between brain and computer/machine by interpreting the thoughts in the brain. BCI can be defined as: "a system that acquires brain signal activity and translates it into an output that can replace, restore, enhance, supplement, or improve the existing brain signal, which can, in turn, modify or change ongoing interactions between the brain and its internal or external environment". Or in simple words, it is a system that translates brain signals into new kinds of outputs as mentioned in \cite{BCI_1}. In intelligent devices, which rely hugely upon the user experience, BCI is used for monitoring purposes, for example monitoring attention level of driver on a long drive. Apart from clinical application for disabled people for controlling devices, BCI is used in virtual reality, gaming industry and entertainment industry. It is also used for meditation and rehabilitation purposes. BCI is employed in boosting learning abilities of children who have difficulty with this.
	
	There are three broad categories of BCIs namely, implantable/invasive, non-invasive and partial/semi-invasive, distinguished by invasively and noninvasively acquired brain signals, respectively. Invasive BCI have a requirement of placing the sensors inside the brain, for which operation is necessary. Although this type of BCI provide one of the best signal-to-noise ratio (SNR), they are not preferred due to high economic cost, short sensor duration and surgical risks. Non-invasive BCI can be realized using brain's electric or magnetic field, out of which electric field is preferred choice as it is mobile and set of apparatus needed for this is easily available and not bulky. Electroencephalography (EEG) has been used to detect the electric signals in a non-invasive manner. BCI using EEG can be further catergorized into event related potential - P300 proposed by \cite{P300_farwell}, steady state visual evoked potential (SSVEP) mentioned in \cite{SSVEP_vidal} and event related desynchronization/synchronization (ERD/S). In this paper, we concentrate on motor imagery system based BCI which is a type of ERD/S.
	
	Motor Imagery based BCI were invented to help people with disabilities to communicate with the outer world. EEG has proven to be effective in motor imagery based BCI due to its very light equipment, low cost and high temporal resolution \cite{Intro1}. However, BCI using EEG suffers from challenges such as extracting features which are useful for specific task due to low specificity, vulnerable to volume conduction effects, non-stationarity, and prone to noise \cite{Intro3}. Another problem posed by EEG signals is that they vary from person to person and session to session \cite{Intro4}. Spatial filtering has been introduced to discriminate between the motor imagery signals using multichannel EEG. The objective behind this filtering is to transform the multichannel EEG signals into small set of channels which are useful to distinguish between the different brain activities \cite{Intro5,Intro6}.
	
	\section{Related Work}\label{related_work}
	
	Common spatial Pattern (CSP) has proven to be very efficient in context of motor imagery based BCI \cite{Intro7,Ramoser,Blankertz,Blankertz_Comp1,Blankertz_Comp3}. Ideally, CSP computes the filters which maximizes the ratio of variance between the brain activities, details of which are explained in the section~\ref{CSP}. However, CSP is vulnerable to noise and problem of overfitting \cite{Reuderink,Grosse}. To overcome these shortcomings, many methods have been proposed in literature \cite{Blankertz,RCSP,CCSP,SCSP}. \cite{Unified} presents detailed comparisons of these regularization techniques. \cite{Dornhege} in their pioneer work extended CSP to multiclass by using methods such as one-versus-rest, pair-wise and simultaneous diagonalization. A major drawback in extending CSP to multiclass is that regularization methods proposed for two class are not effective, therefore accuracies for multiclass are substantially low. For extending original CSP algorithm to more than two class, \cite{Naeem}, \cite{Brunner}, \cite{R1}, \cite{Gouy}, \cite{Gouy1}, \cite{R1} and \cite{hard} proposed different methods using Joint Approximate Diagonalization (JAD) . Among all these approaches common part was that they assumed that CSP is equivalent to finding independent common components for 2-class problem. \cite{Naeem} and \cite{Brunner} researched on different Independent Component Analysis (ICA) algorithms for finding features and components including Informax \cite{Informax}, FastICA \cite{FastICA} and SOBI \cite{SOBI}. They presented key differences between the performance of ICA-based methods with other CSP methods which are variants of One-versus-the- Rest and Pair-Wise \cite{PairWise}. Their findings concluded that overall, CSP methods perform better than ICA-based methods. In addition to this they also stated that among the ICA-based methods,  Infomax achieved better performance than others. CSP is identical to ICA was proven mathematically by \cite{R1}. The authors proposed that CSP by JAD is identical to ICA and also proposed information theoretic approach for feature extraction. Gouy-Pailler in their recent work \cite{Gouy}, \cite{Gouy1} proposed maximum likelihood method for finding spatial filters which is extension of JAD method. They claimed that their method is a neurophysiologically adapted version of JAD. They validated their approach on Dataset 2a of the BCI Competition IV which has nine subjects(they have used only 8 subjects out of 9) with four motor-imagery tasks, and reported that their newly proposed JAD method achieved a better classification accuracy than the CSPs method. They also claimed that CSP with JAD is not better significantly than CSP methods which are variant of  of One-versus-the- Rest and Pair-Wise as reported in the findings of \cite{R1}. In a separate work, \cite{hard} used quadratic optimization to find common spatial patterns for multi-class BCI problems. However, in this case they conducted experiments on their own dataset instead on more widely used datasets so it is difficult to compare their results with previously noted research results. Apart from these two methods there is some interesting work done using subspace method by \cite{Ramoser}, in which they propose Union-based common principal components (UCPC) to create a subspace for class of data from covariance matrix, and finally the union of all the all subspaces is used as common principal component. However, drawback of this method is that chosen principal components may not contribute to some data classes at all.

	In summary, multi-class BCI systems can be broken down into two main groups. The first is based on extension of two class methods that original CSP was designed to work upon, on the other hand other groups deals with JAD. Key point of two class CSP- based methods is that it breaks down the multi-class problems into various classification two-class problems. The two popular methods are One-versus-the-Rest, and a combination of pairs of two-class classification problems (Pair-Wise). Each of these two methods has its own weakness. In the first method there is an assumption that covariance matrix all the other classes are almost identical. However, this assumption hardly stands true in real world data. In latter, there is no surety of getting the CSPs that are optimal of different pairs of classes. This is similar to grouping common principal components of pairs of classes and then finding CSPs. 
	
	Non-stationarity nature of data in EEG presents major problem, which is due to the variation in intra- and inter- session in the same task. This results in very low accuracies in the classification phase, due to poor modeling of data. Recently fuzzy inference systems were used in EEG which has proved to be effective for classification purpose. As shown in \cite{Intro8,type1_coyle,type1_fabien,type1_subasi,type1_yang} type-1 inference systems have been employed for classification of EEG signals. However, type-1 fuzzy sets are inefficient in handling non-stationarity of EEG signals as pointed out by \cite{type1_limit_herman}. Type-2 fuzzy sets which was proposed by \cite{type2_zadeh} along with uncertainty framework can be used to handle non-stationarity as shown in \cite{type2_mendel}. Type-2 sets are computationally very expensive which limits their usage. Modification in type-2 set was presented by \cite{type2_liang}, where uncertainty is represented as bounded interval is highly efficient in computation. Interval type-2 neuro fuzzy inference systems were proposed for classification as mentioned by \cite{type1_limit_herman,IT2NFIS_herman,IT2NFIS_nguyen}. Although they gave decent accuracies, fixed structure of these network have proven to be limitations in handling time varying data of EEG as pointed out by \cite{fixed_lughofer,fixed_lughofer2015,fixed_pratama}. To handle this, evolving structure and learning algorithm was proposed in literature. Self-Regulated Interval Type-2 Neuro-Fuzzy Inference System (SRIT2NFIS) proposed in \cite{Main3} is one such algorithm where structure is evolved based on data. SRIT2NFIS has been used in this paper for classifying the motor imagery EEG details of which are presented in section~\ref{Self}.
	
	\section{Feature Extraction}
	\subsection{Common Spatial Pattern} \label{CSP}
	CSP algorithm is widely used in BCI field and is used to find spatial filters which can maximize variance between classes on which it is conditioned. As mentioned in section~\ref{related_work}, many extensions to basic algorithm has been proposed which has made CSP base line model for feature extraction. We present CSP algorithm for two class problem which is then extended to multiclass. We assume that frequency band and the time window of the input EEG signal is known apriori. In addition to this we assume that signal is jointly gaussian with zero mean. These assumptions put no limitations to the further analysis due to following reasons. First, transformation is linear and mean can be subtracted and added at the end. Second BCI based motor imagery systems using EEG data operates in specific frequencies, hence there are no information present in the higher moments and hence there is no loss of information in assuming gaussian distribution. Given an EEG signal extracted from single trail, $ E $ with dimension $ N \times T $ from $ i^{th} $ class, where N is the number of channels in the recording and T is the number of time samples taken for each channel. The corresponding covariance matrix can be defined as 
	
	\begin{equation}
		C_i =  \frac{EE'}{trace(EE')}
	\end{equation}
	
	where $ ' $ is a transpose operator and $ trace() $ is function which calculates sum of diagonal of matrix. For each class, average spatial covariance matrix ($\tilde{C_i}$) is calculated, for two class we have $ \tilde{C_1} $ and $ \tilde{C_2} $. Composite spatial covariance matrix is generated by adding both the averaged matrix.
	The Composite Matrix can be defined as 
	\begin{equation}
		\tilde{C} = \tilde{C_1} + \tilde{C_2}
	\end{equation}
	
	Eigen values ($ \lambda $) and Eigen vectors ($ U $) of the composite matrix are calculated, which can be represented as
	\begin{equation}
		\tilde{C} = U \lambda U'
	\end{equation}
	
	The eigen values are then sorted in descending order and whitening transformation is applied to de-correlate it
	
	\begin{equation}
		P = \sqrt{\lambda^{-1}} U
	\end{equation}
	
	$ C_1 $ and $ C_2 $ are then transformed as 
	
	\begin{equation}
		S_1 = PC_1P' \quad and \quad S_2 = PC_2P'
	\end{equation}
	
	and by simultaneous diagonalization,
	
	\begin{equation}
		S_1 = B\lambda_1B'\quad and \quad S_2 = B\lambda_2B'; \quad \lambda_1 + \lambda_2  = I
	\end{equation}
	
	where $ B $ is the common eigen vector matrix, $ \lambda_1 $ and $ \lambda_2 $ are the eigen values of $ S_1 $ and $ S_2 $ respectively. $ I $ denotes the identity matrix. Addition of both the eigen values ensures that the largest eigen value in one class correspond to lowest eigen value in other class. This helps in the classification while selecting channel which can effectively distinguish between the two class. Spatial Filter $ W $ can be defined as $ W =(BP')' $ and transformed EEG signal can be viewed as 
	
	\begin{equation}
		Z =  WE
	\end{equation}
	
	CSP are the columns of $ W^{-1} $, which is time invariant source of EEG. From this, features are extracted, relatively very small number $ m $ channels suitable for discrimination. First $ m $ and last $ m $ rows of $ Z $ are selected for classification. Selection of $ m $ is done using following method as proposed by \cite{Blankertz}
	\begin{equation}
		score(W_j) =  \frac{med_j^{(1)}}{med_j^{(1)} + med_j^{(2)} + med_j^{(3)} + med_j^{(4)}}
	\end{equation}
	where,
	
	\begin{center}
		$ med_j^{(c)} = median_{i \in \iota_c} \big(W_j'E_iE_i'W_j \big) $ \hspace*{1mm} $ (c \in {1,2,3,4}) $
	\end{center}
	
	$ \iota $ in a set of trials belonging to a particular class. Ratio of median score near to 1 or 0 indicates good score of the corresponding spatial filter. $ m = 3$ gives optimal discriminable features. Features are selected using value of $ m $ as presented,
	
	\begin{equation}
		f_p  = \log{\left(\frac{var(Z_p)}{\sum\limits_{i=1}^{2m}var(Z_{i})}\right)}; \quad where \quad p = 1,2...,2m 
	\end{equation}
	$ f_1,f_2...f_{2m} $ are the features used for classification. Log transformation projects the data into normal distribution.
	\subsection{Proposed CSP}
	
	To mitigate the effect of noisy trials on eigen values and eigen vectors, the calculation of average covariance matrix for each class, Frobenius norm as presented in \cite{F_norm} of each covariance matrix obtain from different trials is calculated. Frobenius norm is square root of the sum of the diagonal elements of the matrix. The covariance matrices from the same class have a similar magnitude of the Frobenius norm. Z-score \cite{z_score} is calculated for all trials, which transforms Frobenius norm obtained earlier into zero mean and unity standard variance. If the Z-score is more than threshold then that trial is termed as an outlier and is removed from the calculation of CSP filter.
	
	As mentioned in section~\ref{related_work}, JAD based methods were performing better relative to other methods for extending CSP to multi-class. The notion behind JAD is 
	
	\begin{equation}
		W'\tilde{C_i}W = D_i
	\end{equation}
	
	where $ \tilde{C_i} $ is the average covariance matrix of $ i^{th} $ class, W is the spatial filter and $ D_i $ is the corresponding diagonal matrix for each class. The objective is to find W such that, all the classes can be diagonalized as mentioned above. There are many ways of obtaining spatial filters by JAD as stated in section~\ref{related_work}. In this paper we have used method proposed by \cite{JAD_Calc} which uses fast Frobenius diagonalization (FFDIAG). Instead of using log variance as measure of feature extraction as we do in basic CSP, we will be using information theoretic feature extraction \cite{R1} for selecting the relevant spatial filters from $ W $. Information Theoretic Feature Extraction has recently received considerable attention in the machine learning community, mostly in a nonparametric setting. It can be briefly explained as follows; Suppose we have a random variable $ \vec{x} \in \chi$, which is in our application EEG data. Also we suppose that it belongs to one of the class $ c \in C $, where $ C $ is a set of all the classes present in the data. Objective of algorithm is to identify the transformation, which can be mathematically stated as $ f* : \chi \rightarrow \hat{\chi}$, where $ \chi $ is our input feature space and $ \hat{\chi} $ is the discrete set. This transformation should also preserve the information related to the class labels.  
	
	\begin{equation}
		f* = argmax_{f \in \mathbb{F}}\{ I(˘c,f(\vec{x}))\},
	\end{equation}
	
	where this equation, calculates the I(.) in other function space which is mutual information between c and f(.). The basic notion of this algorithm is to provide upper bounds as well as lower bounds on the classification error. Minimum classification is taken into consideration also with respect to mutual information. This bounds are provided by two inequalities which are as follows,
	
	\begin{itemize}
		\item {\textbf{Fano's inequality :-} This inequality provides the lower bound for classifier $ g : \chi \rightarrow C $ and is shown mathematically as 
			\begin{equation}
				P_e := argmin_{g \in G}\{P_r \{c \neq g (f(\vec{x}))\} \geq \frac{H(c|f(\vec{x}))-1}{log|C|} ,
				\\
				= \frac{H(c) - I(c,f(\vec{x}))-1}{log|C|} ,
			\end{equation}
			where H(.) represents Shannon Entropy, $ |C| $ is total number of classes present in $ C $, and $ G $ the set of all classifiers}
		
		\item {\textbf{Upper Bound inequality :-} 
			\begin{equation}
				P_e \leq 1 - 2{I(c,f(\vec{x}))-H(c)} .
			\end{equation}
		}
	\end{itemize}
	
	These two inequalities helps in maximizing the mutual information and reducing the error. In context of BCI, out main objective is to find a reduction in dimension of the input data while simultaneously maximizing the mutual information between the features extracted and their corresponding class labels. 
	\section{Classification} \label{Self}
\begin{figure}[h!]
	\includegraphics[width = 0.5\textwidth]{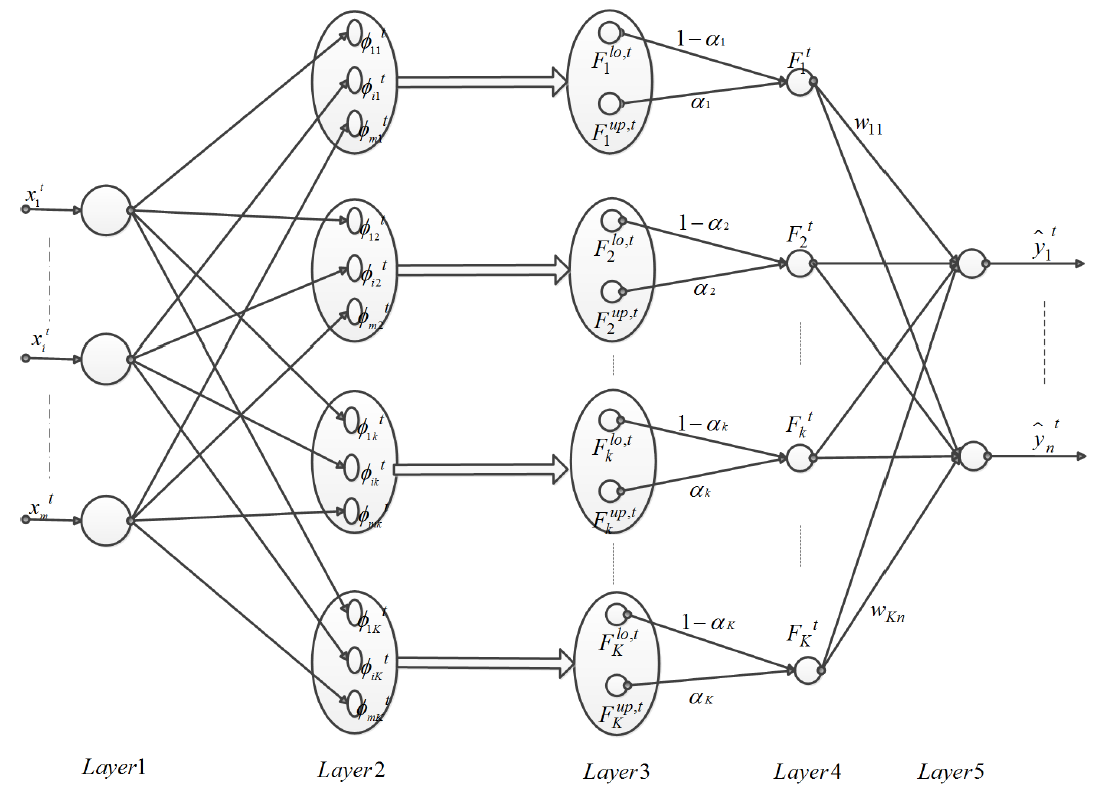}
	\centering
	\caption{Structure of SRIT2NFIS \cite{Main3}}
	\label{fig:Struct_SRT}
\end{figure}

	Extracted features after CSP, are given as the input to the classifier. Keeping in mind the non-stationarity nature of EEG signals and small amount of training data, SRT2NFIS is used for classification. SRIT2NFIS is Takagi-Sugeno-Kang (TSK) type neuro-fuzzy inference system with sequential learning. Uncertainty framework helps in modeling the non-stationarity in the signals. Uncertainty is modeled as Gaussian membership function with uncertain mean and fixed variance. TSK type neuro-fuzzy inference system is embedded upon neural network, these network then can be realised in four to six layers. SRIT2NFIS is a 5 layer network as shown in figure~\ref{fig:Struct_SRT} with projection based learning algorithm. Let us assume that we have a labeled input data, $ (x^t , c^t) $, where $ x^t $ is $ m $-dimensional input vector of $ t^{th} $ sample and $ c^t $ is its corresponding class label. $ c^t \in (1,2,...,n)$ where $ n $ is the total number of class. Output $ y^t $ is described as,
	\begin{equation}
		y_{j}^t = 
		\begin{cases} 
			\quad 1, & if j == c^t \\
			\quad -1, & otherwise 
		\end{cases}
		j = 1,2,...,n
	\end{equation}

The main goal of SRIT2NFIS is to estimate the function which can map input($ x^t $) to output($ y^t $), where $ x^t $ is a $ m $ dimensional input vector and $ y^t $ is the output class label.
\\
\textbf{Layer 1 - Input Layer: }Input layer directly maps the input to the layer 2, which can be represented as $ u^t_i = x^t_i $ for $ t^{th} $ sample and $ i  = 1,2,...,m$\\
\textbf{Layer 2 - Fuzzification Layer: }In this layer, type-2 Gaussian membership function is used in each of the neuron to model the non-stationarity in the input.\\
\textbf{Layer 3 - Firing Layer: }This layer calculates the firing strength which is represented by each neuron of this layer. Algebraic product of the membership values amounts to the strength, output of this layer is type-1 fuzzy set.\\
\textbf{Layer 4 - Type reduction layer: }Using Nie-Tan type-reduction approach \cite{archi_1}, each neuron converts the input type-1 fuzzy set to a fuzzy number for each rule. $ \alpha $ is set to 0.5, which is the weighted measure of uncertainty.\\
\textbf{Layer 5 - Output layer: }Normalized weighted sum of the firing strength is calculated in this layer, output of which gives the predicted output class of each trail.

It uses a projection based algorithm \cite{PBL}, with minimization of sum of squared hinge error for avoiding over-fitting where the regularization parameter is set to 0.01. When a new sample arrives at the input of the network, prediction error and spherical potential \cite{Spherical} are calculated. Based on these two parameters, decision is taken either to keep the sample and modify parameter/addition of rule, discard the sample or reserve it for the later use. The learning algorithm parameters are discussed as follows:

\textit{Absolute maximum hinge-error:} It is calculated to measure the prediction error of the current sample.

\textit{Class-Specific Spherical potential:} Knowledge content in the sample is measured using spherical potential for that particular class. It is calculated by averaging the distance in the hyper dimensional space between current sample and the rules from the same class which are contributing significantly. Higher the potential lower is the novelty in the sample.

\textit{Rule growing criterion:} If the novelty of the current sample is high and absolute maximum hinge-error is also high, new rule is added in the network provided the predicted class label is incorrect or no rule has been fired for the current sample. This is controlled by two threshold; Add threshold and novelty threshold. Add threshold is a adaptive in nature, main objective of this is to capture global knowledge at the start and then fine tune accordingly. It's adaptiveness is controlled by another parameter $ \gamma $ which is set to 0.99, this value has been estimated due to small training samples. Range for add threshold is [1.01, 1.20] and that for novelty threshold is [0.01, 0.60].

\textit{Intra/Inter class overlap factor :} Width of the new rule is influenced by the nearest intra/inter class rules. Intra class overlap factor should be higher as we need a higher overlap in case of BCI, it is set to 0.95. However, Inter class overlap factor is chosen in range [0.1, 0.4], very little overlap will hinder the generalization capability of network.

\textit{Parameter update criterion:} Adaptation of the output network weights are carried out whenever the predicted class label is correct and absolute hinge error is greater than update threshold. Update threshold is also a adaptive parameter and it is also controlled by $ \gamma $ mentioned earlier. Rule add threshold range is [0.04, 0.2].

\textit{Rule Pruning Criterion:} If for consequent number of samples greater than pruning window of the same class, contribution of the rule is not significant then that rule is removed from the network. Pruning threshold is set to 0.01 and pruning window is set to 10.

\textit{Sample deletion criterion:} If current sample contains no novelty then that sample is discarded, sample delete threshold is set to 0.05.

\textit{Sample reserve criterion:} If the sample does not meet any of the above mentioned criteria, then it is reserved for the later use by pushing this sample at last of the sequence.

\section{Results and Discussion}
\subsection{Data}
\begin{figure}[h!]
	\centering
	\includegraphics[width=0.7\textwidth]{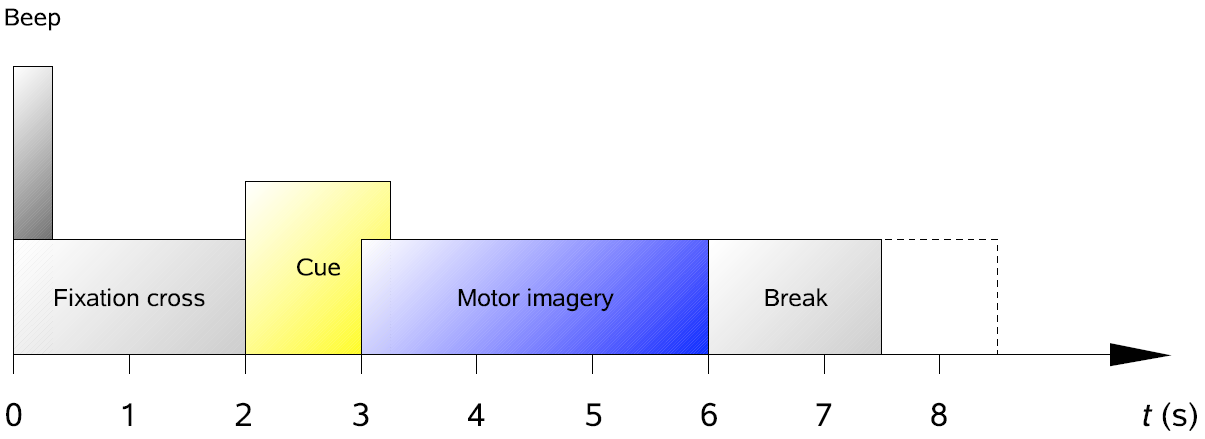}
	
	\caption{Timing scheme of Data set 2 \cite{dataset}}
	\label{fig:Time_data2}
\end{figure}

We have implemented proposed method on benchmark dataset of BCI competition IV namely dataset 2a. It is a continuous Multiclass Motor Imagery EEG data of 9 subjects. It consisted of four different motor imagery class, imagination of the left hand, right hand, both feet and tongue. Data was recorded in two session on different days for each patient. Each sessions consisted of 6 runs and each run comprised of 48 trials (12 for each class). So we have 288 trial for each session. Each trail consisted of 7.5 sec recording. At the beginning of each trail a fixation cross appeared on the screen and also an acoustic warning tone was also presented. After 2 sec cue was presented on screen in the form of arrow which pointed either left, right, down or up corresponding to each of the four class. Subjects were asked to imagine respective movement after the cue has been presented till sixth sec. This was followed by a 1.5 sec break. Figure~\ref{fig:Time_data2} shows the graphical view of the paradigm used for recording. 

The signals were samples at 250Hz and then filtered using a band pass filter having 0.5Hz as low and 100Hz as high frequency. Along with this it was also filtered using 50Hz notch filter for removing the line noise. This experiment was done using 22 electrodes. For each subject training and testing data set was provided separately. 

\subsection{Experiment}

We have used training data to train our network and results are generated for the testing data. For evaluation metric overall classification is used as a parameter for comparison. Accuracy is defined as ratio of correctly classified trials in their respective class to total number of trials for that subject.

%

\begin{table}[]
	\caption{Comparision with Other Approaches (Accuracy in \%)}
	\resizebox{\textwidth}{!}{%
		\begin{tabular}{|lcccc|}
			\hline
			\multicolumn{1}{|c}{Subjects} & \multicolumn{1}{c}{\begin{tabular}[c]{@{}c@{}}Multiclass\\CSP (mCSP)\\~\cite{R1}\end{tabular}} & \multicolumn{1}{c}{\begin{tabular}[c]{@{}c@{}}ComplexCSP\\~\cite{R2}\end{tabular}} & \multicolumn{1}{c}{\begin{tabular}[c]{@{}c@{}}Proposed CSP with\\ SVM\end{tabular}} & \multicolumn{1}{c|}{\begin{tabular}[c]{@{}c@{}}Proposed CSP with\\ SRIT2NFIS\end{tabular}} \\ \hline
			\multicolumn{1}{|l|}{1} &  48.1 & 61.5 & 68.75 & \textbf{74.65}  \\
			\multicolumn{1}{|l|}{2} &  27.3 & 32.1 & 41.67 & \textbf{45.48}  \\
			\multicolumn{1}{|l|}{3} &  70.6 & 68.6 & 66.31 & \textbf{74.31}  \\
			\multicolumn{1}{|l|}{4} &  21.4 & 27.1 & 37.98 & \textbf{39.58}  \\
			\multicolumn{1}{|l|}{5} &  22.7 & \textbf{34.3} & 25 & 32.99  \\
			\multicolumn{1}{|l|}{6} &  32.4 & 35.3 & 36.62 & \textbf{37.9}  \\
			\multicolumn{1}{|l|}{7} &  52.3 & 48  & 52.97 & \textbf{54.17}  \\
			\multicolumn{1}{|l|}{8} &  65.8 & 65.6 & 65.55 & \textbf{66.32}  \\
			\multicolumn{1}{|l|}{9} &  34.2 & 41.8 & 64.58 & \textbf{66.31}  \\
			\multicolumn{1}{|l|}{Mean} &  41.64 & 46.01 & 51.04 & \textbf{54.63}  \\
			\multicolumn{1}{|l|}{SD} &  18.34 & \textbf{15.65} & 16.15 & 16.27  \\ \hline
		\end{tabular}%
	}
\label{table:MultiClass}
\end{table}

\begin{figure}[h!]
	\centering
	\includegraphics[width=0.45\textwidth]{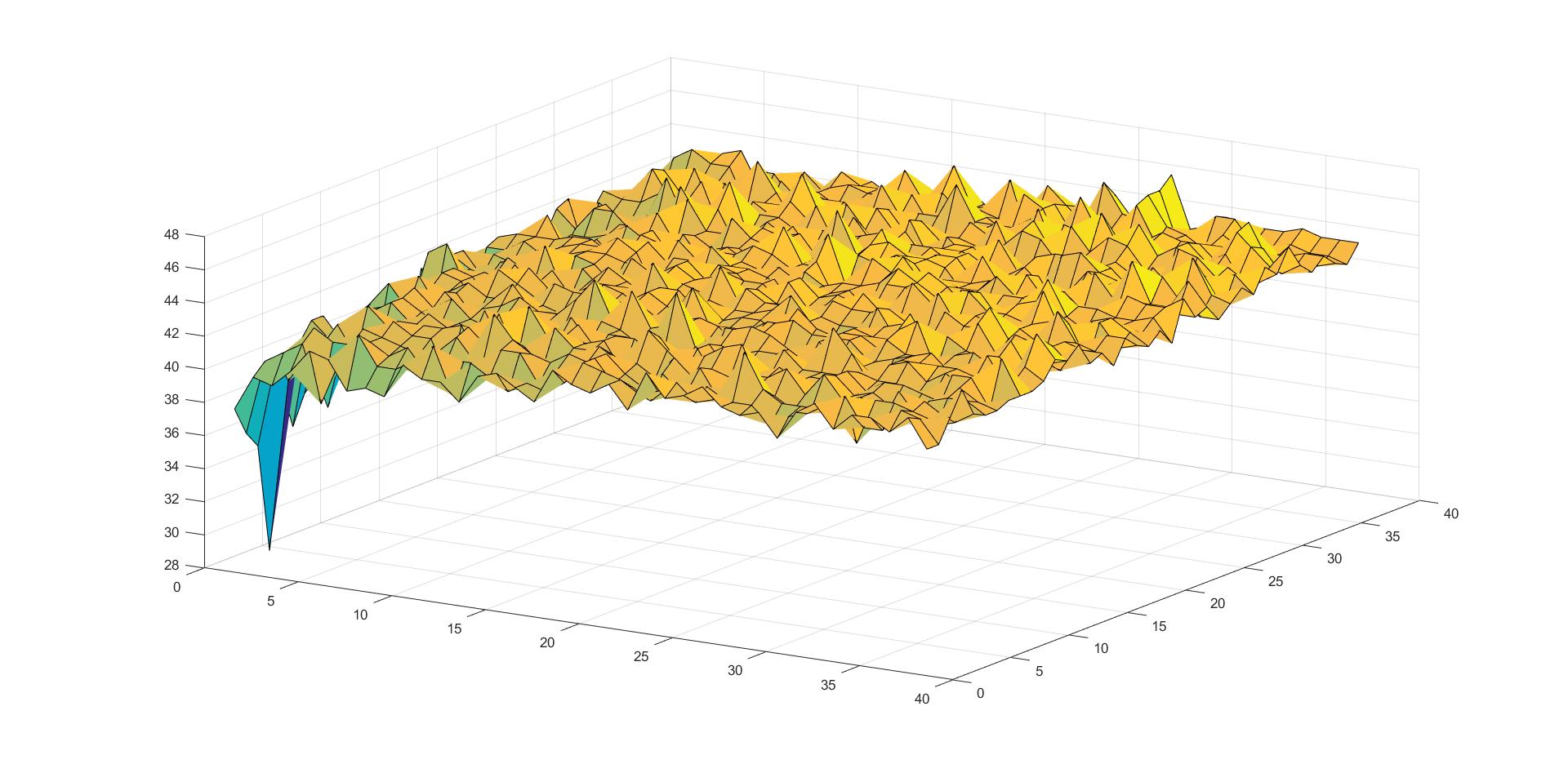}
	
	\caption{Surf plot for Parameter Width Vs Iteration vs Accuracy}
	\label{fig:Surf_plot}
\end{figure}

Data is preprocessed before feeding to proposed mechanism. Butterworth Band pass filter of order 5 used to filter signals between 8-40Hz as proposed in \cite{filter}. Proposed CSP method was used to extract features with $ m = 3$, as described in section~\ref{CSP}. While using SRIT2NFIS as classifier, parameters such as add threshold, update threshold, novelty and inter-class overlap have to be optimized for each patient. This was achieved by implementing particle swarm optimization (PSO) for each subject independently. Two parameters that govern the PSO are number of iteration and parameter width. Figure~\ref{fig:Surf_plot} shows the plot of these two parameters and accuracy for subject 2. The two parameters differ from subject to subject. Best accuracy for subject 2 was 45.48\% which is 66.59\% increase in accuracy reported in \cite{R1}. Addition to this, it is interesting to note here that subject 2 and subject 5 have a very noisy data as compared to other subjects which is evident from the earlier results in the literature and also this can be seen in the two class classification results on same data.

Table~\ref{table:MultiClass} shows the comparison of proposed approach and Multiclass CSP (mCSP) \cite{R1} and ComplexCSP \cite{R2}. Comparison of proposed approach is done with Multiclass CSP (mCSP) \cite{R1} and ComplexCSP \cite{R2}, both of these algorithms use Support vector Machine (SVM) as their classifier. The increase in accuracy can be attributed to two major arguments; formation of generalized eigen vectors due to removal of noisy trials which causes a better feature after CSP and better handling of over fitting and non stationarity of EEG data by SRIT2NFIS by its self regulating learning algorithm.

\section{Conclusion}
In this paper we have presented common spatial pattern algorithm for multiclass EEG classification with a pre-processing step which can improve the generalization of CSP covariance matrices by removing the trials which are noisy/affected with artefact. SRIT2NFIS is used as a classifier to handle the non stationarity present in the signal. The results were presented on publically available data set BCI competition IV data 2a. Results are compared with the currently state of the art algorithms for multiclass classification. It clearly shows that proposed method outperforms the in four class classification by improving the mean accuracy 8-13\%. The increase is significant in subjects which earlier had very low accuracies.

\section*{ACKNOWLEDGMENT}

Authors would like to thank Prof. N. Sundararajan, Prof. S. Suresh and Mr. Ankit Das for their immense support in this research work.

\bibliographystyle{ACM-Reference-Format}
\bibliography{Bibliography}

\end{document}